\documentclass[letterpaper, 10 pt, conference]{ieeeconf}  

\IEEEoverridecommandlockouts                              

\usepackage{graphics} 
\usepackage{epsfig} 
\usepackage{mathptmx} 
\usepackage{times} 
\usepackage{amsmath} 
\usepackage{amssymb}  

\usepackage{xcolor}

\title{\LARGE \bf
Planning to Build Soma Blocks Using a Dual-arm Robot}

\author{Hao Chen, Weiwei Wan, Keisuke Koyama, and Kensuke Harada
\thanks{All authors are affiliated with the Graduate School of Engineering Science, Osaka University, Japan.}
\thanks{Contact: Weiwei Wan, {\tt\small wan@sys.es.osaka-u.ac.jp}}%
}

\begin{document}

\maketitle
\thispagestyle{empty}
\pagestyle{empty}

\begin{abstract}
This paper presents a planner that can automatically find an optimal assembly sequence for a dual-arm robot to assemble the soma blocks. The planner uses the mesh model of objects and the final state of the assembly to generate all possible assembly sequence and evaluate the optimal assembly sequence by considering the stability, graspability, assemblability, as well as the need for a second arm. Especially, the need for a second arm is considered when supports from worktables and other workpieces are not enough to produce a stable assembly. The planner will refer to an assisting grasp to additionally hold and support the unstable components so that the robot can further assemble new workpieces and finally reach a stable state. The output of the planner is the optimal assembly orders, candidate grasps, assembly directions, and the assisting grasps if any. The output of the planner can be used to guide a dual-arm robot to perform the assembly task. The planner is verified in both simulations and real-world executions.
\end{abstract}

\section{INTRODUCTION}
Deploying robots to assemble workpieces in modern manufacturing is a tough job. It requires many skilled system engineers to design fixtures, end-effectors, assembly sequences, as well as motion sequences to enable industrial robots to perform a given assembly task.  The process is very time-consuming and error-prone.  In this paper, we present a planner that can automatically find an optimal assembly sequence for a dual-arm robot to assemble objects. The planner is developed for dual-arm robots since two arms are more flexible and reduce the need to design new fixtures and end-effectors. The planner optimizes the dual-arm assembly sequences automatically and reduces the need to design each assembly and motion sequence. It is expected to significantly reduce the cost and minimizes the manufacturing time and manpower.

\begin{figure}[t]
\begin{center}
\includegraphics[width=\linewidth]{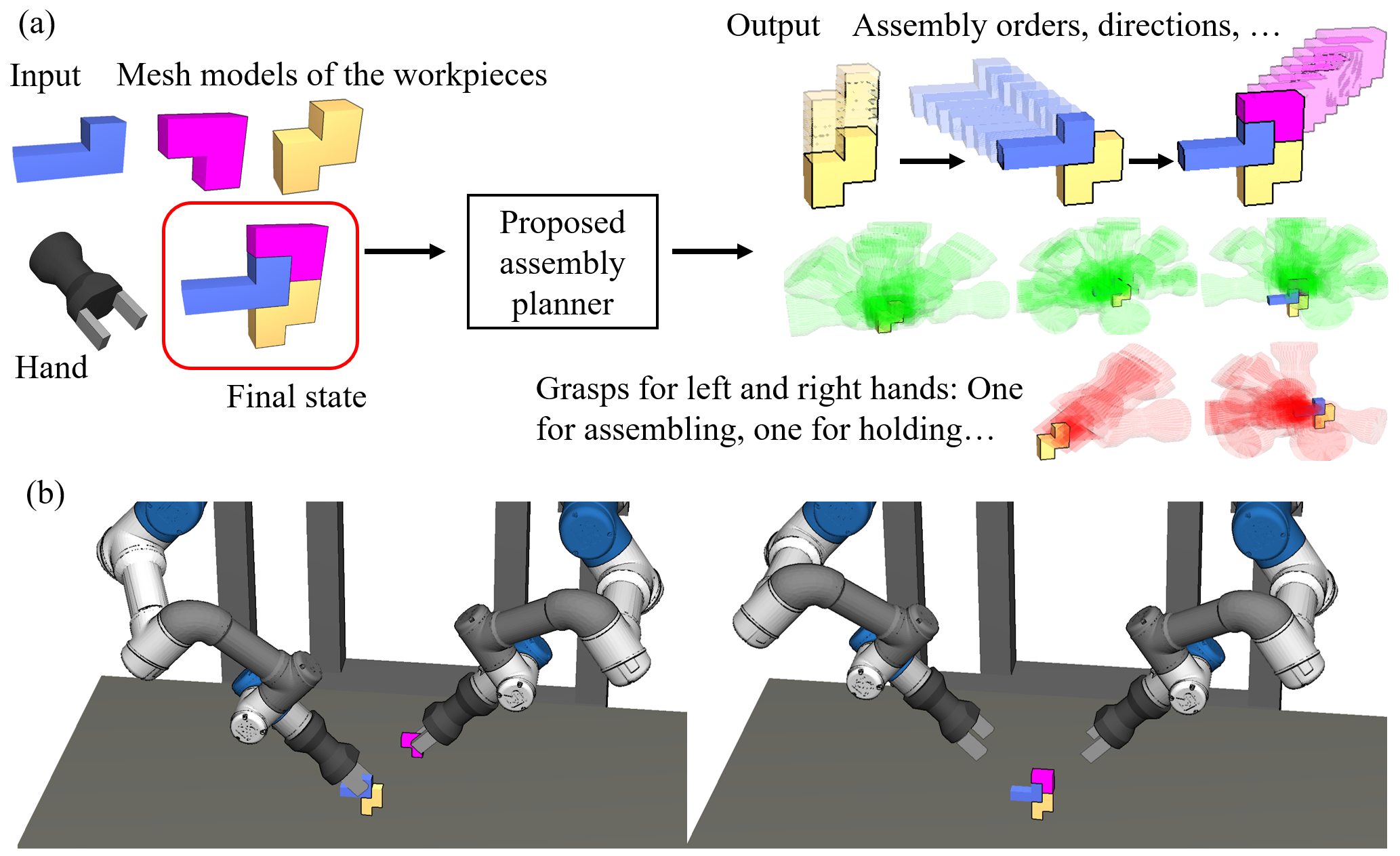}
\caption{(a) The input and output of the proposed planner. (b) A dual-arm robot assembles the soma block by using its right hand to play the role of the assisting grasp found by our planner. In the left subfigure, the right hand is holding and supporting the two unstable workpieces in the assembly.}
\label{fig:intro}
\end{center}
\end{figure}

In our previous work, we develop an assembly planner that can automatically plan an optimal sequence for assembling objects using a single robot arm \cite{wan2018assembly}. The output of the planner can be readily used by a motion planner to produce robotic assembly motions. One of the drawbacks in our previous work is that it assumes that the whole assembly process is stable, which may lessen the number of solutions in the real situation. For example, when the human workers do the assembly work, they often use one hand to stabilize the unstable components and use the other hand to assemble the mating workpiece. For the assembly process where the unstable components exist, like the assembly shown in the red circle of Fig.\ref{fig:intro}(a), the previous planner cannot find a solution. The assembly will fall apart in the middle if not well supported. Note that the unstable components we mentioned here mean that the current state is unstable or adding a new component in some future steps will result in an unstable state. The final assembly state must be stable otherwise the assembly planning is meaningless.
 
Similar to the previous work, the planner proposed in this paper permutates, evaluates, and searches the assembly sequences by considering the stability, graspability, and assemblability. The main improvement lies in that we develop a more robust stability analyzer and propose a method that refers to a second arm to handle the unstable components. The proposed planner first permutates the workpieces involved in the assembly and generates all possible assembly sequences. Then, it analyzes every permutation sequentially by considering the stability, graspability, and assemblability. During this process, the planner can find the accessible grasps of the workpieces that are collision-free, force-closed, and have an assembly direction that is robust to uncertainty. After that, the planner detects if adding a new component results in instability, the planner will use one of the dual-arm robot hands as external assistance to hold and support the unstable components. Fig.\ref{fig:intro} illustrates our planner. The input and the output of the planner are shown in (a). The input to the planner includes
\begin{itemize}
\item The mesh model of the robotic hand;
\item The mesh models of the objects;
\item The final state of the assembly.
\end{itemize}
The output includes 
\begin{itemize}
    \item The assembly order;
    \item The accessible grasp;
    \item The assembly direction;
    \item The assisting grasps (if unstable components exist).
\end{itemize}
Note that we assume a dual-arm robot in the study. Nevertheless, the planner is not limited to two arms. A dual-arm robot can provide one assisting hand to support the unstable components. Multiple arms could provide hands to support more components.  

The main contribution of this work is using dual-arm manipulation to stabilize the assembly by holding and supporting the unstable components. To our best effort, we didn't find a similar assembly planner that considers the dual-arm manipulation for stabilization in contemporary studies. The output of our proposed planner can be directly used to guide a dual-arm robot to perform the assembly task. A motion planner can be integrated afterward to produce the assembly motions to control each robotic arm.

The paper is organized as follows: Section II presents the related studies. Section III shows the overall workflow of the proposed planner. Section IV presents the details of each method in the workflow. Section V is the experimental section. It analyzes the result of the proposed planner with both simulations and real-world executions. It also demonstrates the integration of the proposed assembly sequence planner with some motion planners. The final section concludes the paper and presents future work.

\section{RELATED WORK}
Numerous experimental and theoretical studies have been performing to assembly sequence planning. The early researches mainly considered the geometry constraints on the CAD model or involved the human experts as an instructor to plan the assembly sequence. The early research done by Mello et al. \cite{de1991correct} defined the assemblies by using logical expressions and generate the assembly sequences from AND/OR graphical method. Baldwin et al. implemented an interactive assembly sequence planning software to assist the human experts to find the optimal sequence by providing criteria like stability, fixturing, and orientation. Dini et al. \cite{dini1992automated} proposed an approach that could detect the subassemblies by inputting the interference, contact, and connection matrices of the product. Wilson et al. \cite{wilson1994geometric} proposed an approach to find the assembly sequence by building a non-directional blocking graph (NDBG) from geometrics constraints and reasoning about the graph. Ritchie et al. \cite{ritchie1999generation} proposed the method to generate the assembly sequence by letting humans using the immersive virtual reality and record human behavior as a reference to generate the assembly sequence. Halperin et al. \cite{halperin2000general} presented the motion space that was derived from the configuration space of motion planning to build an NDBG for finding assembly sequence. Mok el at. \cite{mok2001automatic} designed an algorithm for finding the assembling or disassembly sequence by considering the topological constraints from the CAD model and represented the sequence by using a structured assembly coding system (SACS). Jimenez el at. \cite{jimenez2013survey} made a survey investigating the geometric reasoning used for assembly sequence planning.

The early assembly planning systems considered only geometric constraints. More recent work considers a mixed model of constraints. For example, Dobashi et al. \cite{dobashi2014robust} additionally considered the collision-free grasps between manipulated objects and the assembled objects during assembling, although the assembly sequence is pre-defined manually considering these constraints. A research done by Dogar et al. \cite{dogar2019multi} found the assembly sequence for several mobile robots to assemble a chair. The method considered the constraints between different robots, between the robot and assembly components, and between mating components and assembled components. Rodriguez et al. \cite{rodriguez2019iteratively} proposed an assembly planner that checked the geometric constraints and collision of the robot gripper in the high level and checked the assembly sequence by using simulation robotic manipulation in the low level only if constraints in the high level are satisfied. Tian et al. \cite{tian2017disassembly} considered the component uncertainty quality and varying operational costs in the disassembly in assembly sequence planning. Li et al.\cite{li2019sequence} and Aliev et al. \cite{aliev2019task} respectively considered the human-robot collaboration situation when determining the assembly sequence.

Compared with these studies, our proposed planner considers the advantage of dual-arm manipulation in planning the assembly sequence. The planner can refer to an assisting grasp from a second arm to additionally hold and support the unstable assembly so that the first arm can further assemble new workpieces and finally reach a stable state.

\section{OVERALL DESIGN}
\begin{figure}[!htbp]
  \begin{center}
  \includegraphics[width=\linewidth]{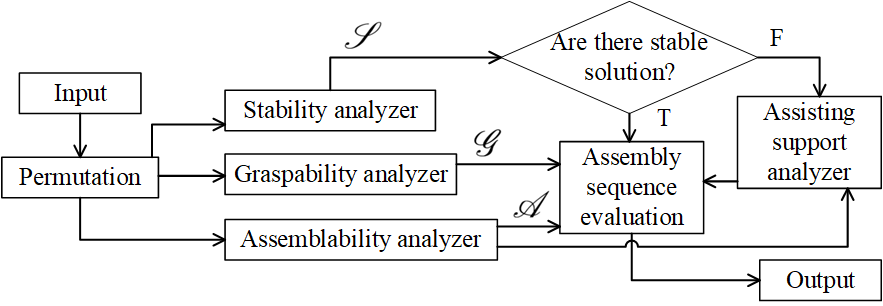}
  \caption{The flowchart of the proposed planner.}
  \label{fig:flowchart}
  \end{center}
\end{figure}

The overall flowchart of the proposed assembly planner is shown in Fig.\ref{fig:flowchart}. The input to the planner includes: 1) The mesh model of every workpiece for assembly; 2) The final assembly pose of every workpiece. The output includes: 1) The assembly order of each workpiece; 2) The assembly direction for each workpiece; 3) The grasp of each workpiece; 4) The assisting grasp for assembly (if unstable components exist). By analyzing the geometric constraints between the assembled components and the upcoming workpieces, the proposed planner finds an index to represent the stability, graspability and assemblability qualities for each assembly permutation. Then, by comparing the index for each assembly permutation, the planner finds the optimal assembly sequence. The planner can also get the assembly direction, the grasps for each workpiece, and the assisting grasps for stabilizing unbalanced force and torque during the optimization. A motion planner (e.g. \cite{wan2016achieving}) can use this output to plan the motions for a dual-arm robot to conduct the assembly task.

A detailed description of Fig.\ref{fig:flowchart} is as follows.
The ``Permutation'' box in the left computes all possible assembly orders. The orders are evaluated by the stability, graspability, and assemblability analyzers in the middle. The stability analyzer reckons the stability quality $\mathcal{S}$ of the subassembly by evaluating the frictional forces at the contact points of the workpieces touching each other and measuring a component's stability by its ability to resist the disturbance (A component is a cluster of assembled workpieces). The ``Stability analyzer'' output the stability quality. If all the assembly orders are unstable, namely the stability quality is 0, the ``Assisting support analyzer'' will be triggered to find assisting grasps to hold and support the unstable component. The assisting grasp is from the second arm. It is expected to make the unstable component stable. The ``Assisting support analyzer'' will recompute the new stability quality for the assembly order after being triggered. The "Assisting support analyzer" outputs the updated stability quality as well as the assisting grasps.
The ``Graspability analyzer'' computes the graspability quality $\mathcal{G}$ for each workpiece by computing the number of the force-closed and collision-free grasps. The computed grasps information can be sent to the motion planner in a later stage for a robot to back chain the picking and moving of the workpiece. The ``Graspability analyzer'' outputs the graspability quality and accessible grasps.
The ``Assemblability analyzer'' determines the assemblability quality $\mathcal{A}$ by evaluating the normals of the contact faces between the assembled components and the upcoming workpieces. The assembly direction can be found in this process by considering the direction that has the largest clearance from contact normals. The "Assemblability analyzer" outputs the assemblability quality and assembly directions.
Finally, the ``Assembly sequence evaluation'' box measures an optimal sequence by considering $max(min(\mathcal{S}),min(\mathcal{G}),min(\mathcal{A}))$. It outputs the assembly order, the grasps, and the assembly directions of the chosen optimal order. Also, it will toggle on the output of the ``Assisting support analyzer'' to get the assisting grasps (if exists).
The detail will be explained in the following section.

\section{IMPLEMENTED DETAILS}
\subsection{Permutation} 
The permutation is to generate all possible assembly orders for the consequential analyzers to find the optimal assembly order. The entire number of potential assembly order of a assembly with n workpieces is calculated by $n!$. For example, an assembly with three workpieces ``A", ``B", ``C" has $3!=6$ possible assembly orders. They are
A $\leftarrow$ B $\leftarrow$ C, 
A $\leftarrow$ C $\leftarrow$ B, 
B $\leftarrow$ A $\leftarrow$ C, 
B $\leftarrow$ C $\leftarrow$ A, 
C $\leftarrow$ A $\leftarrow$ B, 
C $\leftarrow$ B $\leftarrow$ A. 
Here, A $\leftarrow$ B $\leftarrow$ C indicates that assemble A to B first and then assemble C to the complex of A and B. When assembling the B to the A, we define that the B is an upcoming workpiece, A is the assembled component, and C is next preparing workpiece. The next preparing workpiece is the following workpiece after finishing assembling the upcoming piece. When assembling the C to the complex of A and B, the C is the upcoming workpiece, A and B is the component, and there is no next preparing workpiece.

Note that at this stage, we do not do any pruning. The goal is to prepare all possible choices of orders for late filtering.

\subsection{Stability analyzer} 
The stability analyzer computes the stability quality of a given assembly order. It sequentially investigates each workpiece in the order. The first step is to check whether the upcoming workpiece is stable after being assembled following a given assembly order. The analyzer evaluates the stability quality of the upcoming workpiece by using the concept of the Grasp Wrench Space (GWS) \cite{strandberg2006method} as a reference. The analyzer finds the contact surface and uses the vertices of the contact surface as the contact points. The static friction model (Coulomb friction) \cite{erdmann1994representation} is used to analyze the contact. We set the local contact coordinate frame at each contact point with the z-axis pointing in the direction of the inward surface normal. The contact force at $i$th contact point is  $\mathbf{f_i}=[f_{x_i}\  f_{{y_i}} \ f_{{z_i}}]^T$. The contact force $\mathbf{f_i}$ should lie in the friction cone $FC_i$ to avoid the slip and separation:
\begin{equation}
     FC_i =\{f_i|\sqrt{f^2_{{x_i}}+f^2_{{y_i}}} \leq \mu_i f_{z_i}\}.
\end{equation}
Here, $\mu_i$ is the coefficient of static friction that depends on the material of the contacting workpieces at the $i$th contact point. We use an inversed 6-side pyramid to approximate the friction cone and balance the trade-off between the complexity and accuracy of the approximation. The wrench is represented by $\mathfrak{\mathbf{w}}=[\mathbf{f}\ \mathbf{\tau}]^T$. Its reference coordinate frame originates at the center of mass of the upcoming workpiece. The z-axis is in the same direction as the world coordinate. Suppose there are k contact points in the contact surface. Since $\mathbf{f_i}$ is bounded in the friction cone $FC_i$, the wrench set $\mathbf{W}$ that can resist all the wrench generated by the sum of all $\mathbf{f_i}$ acting on the upcoming workpiece in the reference coordinate frame can be written as:
\begin{equation}
\mathfrak{\mathbf{W}} = \sum\limits_{i=1}^k \begin{bmatrix}
\mathbf{R_i} & 0\\
p_i\times\mathbf{R_i} & \mathbf{R_i}
\end{bmatrix} FC_i + \mathbf{w_0},
\end{equation}
where $p_i$ is the position of the $i$th contact point in the reference coordinate frame. $\mathbf{R_i}$ is the relative orientation of the $i$th local contact coordinate frame. The $\mathbf{w_0}$ represents the wrench generated by the gravitational force. The stability quality is 0 if the origin of the wrench space is not in the interior of the convex hull of the $\mathbf{W}$. In that case, the upcoming workpiece is unstable and cannot be directly assembled. If the stability quality is not equal to 0, a concrete value will be computed by measuring the shortest distance between the origin in the wrench space and the convex hull of the $\mathbf{W}$. The value indicates the ability to resist external disturbance wrenches. It is represented by:
\begin{equation}
s = min\{d|d \in convexhull({\mathbf{W}})\}.
\end{equation}

\begin{figure}[!htbp]
  \begin{center}
  \includegraphics[width=\linewidth]{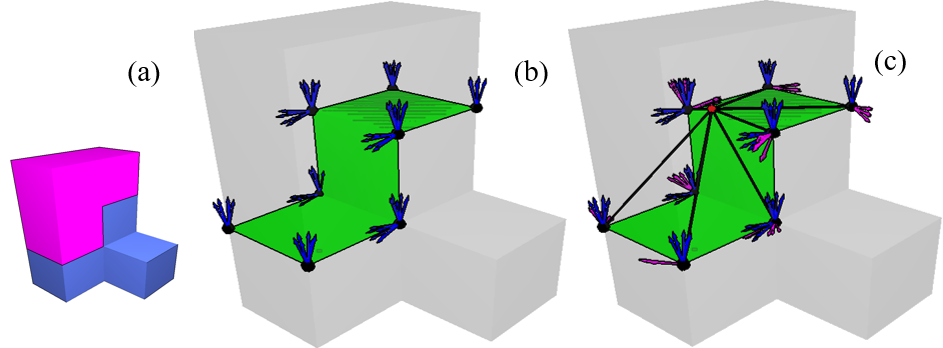}
  \caption{(a) The upcoming workpiece is in pink and the assembled component is in blue. (b) shows the contact surface (green area), the contact points (the black points) and the simplified 6-side pyramid friction cone at each contact point. (c) shows the distance from the center of mass of the upcoming workpiece to each contact point (the black segments), the center of the mass (the red point) and the direction of the torque of every frictional force of the friction cone (the pink arrows).}
  \label{fig:stability}
  \end{center}
\end{figure}

For an assembly with $n$ components, the stability quality of a potential order is  $\mathbf{s}=(s_1 \ s_2 \  \ldots \ s_n)$, where $s_i$ is the stability quality of the $i$th workpiece.
The stability quality of $m$ different assembly orders is thus represented as a matrix like follows:
\begin{equation}\label{eq:stability}
\mathcal{S}=
\begin{pmatrix}
\mathbf{s_1} \\
\mathbf{s_2} \\
\ldots \\
\mathbf{s_m}
\end{pmatrix} 
= 
\begin{pmatrix}
s_{11} & s_{12} & \ldots & s_{1n} \\
s_{21} &s_{22} & \ldots & s_{2n} \\
\ldots & \ldots & \ldots & \ldots \\
s_{m1} & s_{m2} & \ldots & s_{mn} 
\end{pmatrix}
\end{equation}

\subsection{Graspability analyzer} 
\begin{figure}[!htbp]
  \begin{center}
  \includegraphics[width=.9\linewidth]{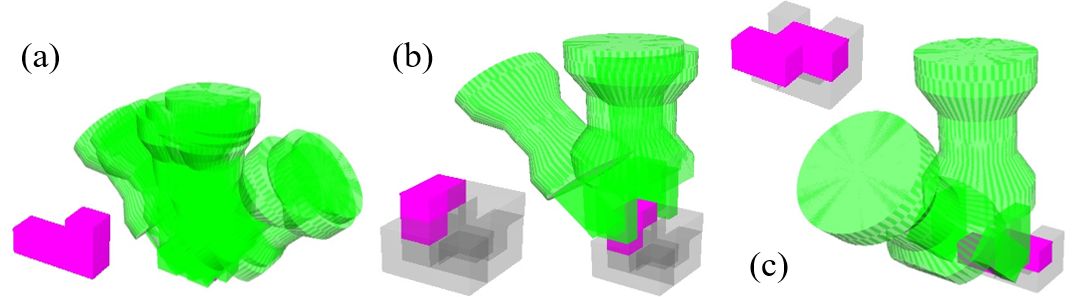}
  \caption{Purple objects: The upcoming workpiece. Gray objects: The assembled components. Green hands: The force-closed and collision-free grasps acting on the upcoming workpiece.}
  \label{fig:grasp}
  \end{center}
\end{figure}

The graspability of a potential order is computed by successively evaluating the force-closed and collision-free grasps of each workpiece. First, we use the method presented in Wan et al. \cite{wan2016achieving} to get the force-closure and collision-free grasps. The workflow of the method is: (i) Find the planar facets;
(ii) Sample the facets; (iii) Find the candidate samples in a parallel surface for attaching a gripper.
Some examples of the successively planned grasps are shown in Fig.\ref{fig:grasp}. The quality of the graspability is evaluated by the number of grasps on the workpiece. For a assembly with $n$ workpieces, the graspability quality of a potential order is $\textbf{g}$ = ($g_1$ $g_2$ \ldots $g_n$), where $g_i$ is the number of grasps for the $i$th workpiece. The graspability quality of $m$ different assembly orders is thus represented as a matrix like:
\begin{equation}\label{eq:graspability}
\mathcal{G}=
\begin{pmatrix}
\mathbf{g_1} \\
\mathbf{g_2} \\
\ldots \\
\mathbf{g_m}
\end{pmatrix} 
= 
\begin{pmatrix}
g_{11} & g_{12} & \ldots & g_{1n} \\
g_{21} & g_{22} & \ldots & g_{2n} \\
\ldots & \ldots & \ldots & \ldots \\
g_{m1} & g_{m2} & \ldots & g_{mn} 
\end{pmatrix}
\end{equation}

\subsection{Assemblability analyzer} 
\begin{figure}[!htbp]
    \centerline{\includegraphics[width=\linewidth]{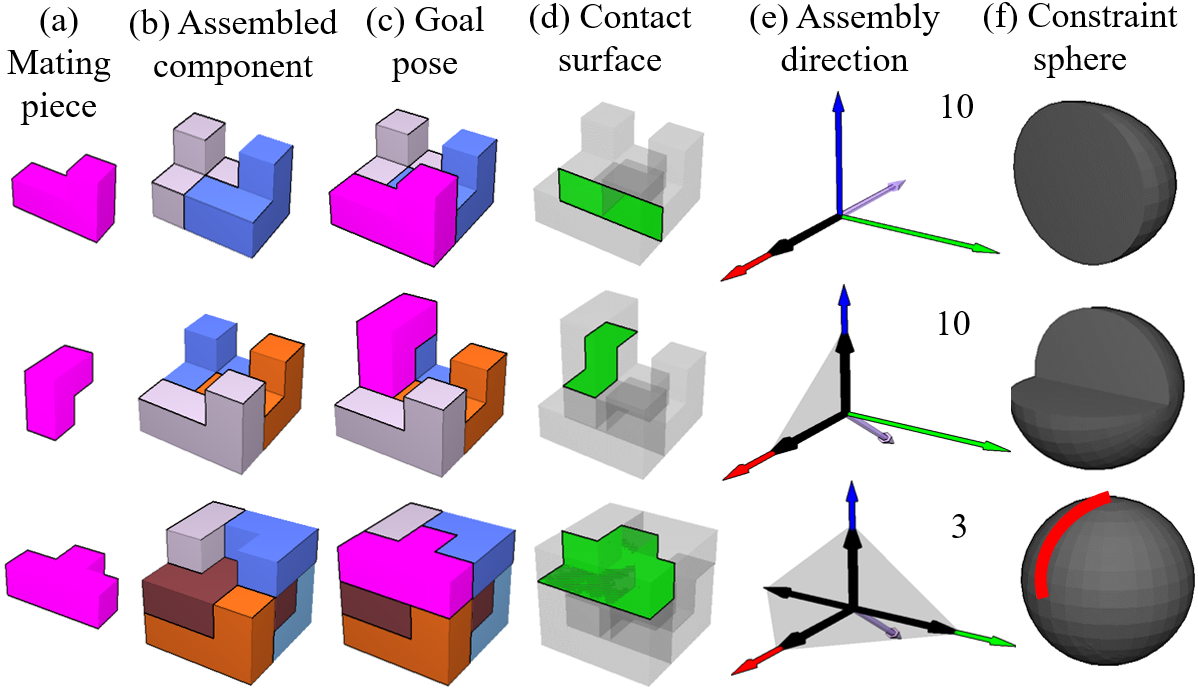}}
    \caption{(a) The upcoming workpiece to be assembled. (b) The assembled components. (c) The goal pose of the upcoming workpiece. (d) The contact surface between the upcoming workpiece and the assembled component. (e) The contact normals (the black arrows), the optimal chosen assembly direction (the purple arrow), and the convex hull (shadow region) rendered in the contact space. The indices at the top-right corners are the assemblability qualities. (f) The correspondent constraint spheres of the contact spaces in (e).}
    \label{fg:assemblability}
\end{figure}

The method we use to compute the assemblability is based on Wan et al. \cite{wan2018assembly}. It classifies an assembly into 9 different types and assigns both the quality index and optimal assembly direction for each type. Fig.\ref{fg:assemblability} shows three examples. The output is the corresponding optimal assembly direction and the assemblability quality for the upcoming workpiece. The assembly directions are constrained by the black arrows in column (e). The optimal one is denoted by the purple arrow, which has the largest clearance to all constraining black arrows. The purple arrow is perpendicular to the y-axis and points inversely against the shadow region. The assembly quality is denoted by the number on the top right corner of each small figure in column (e). These values are computed by analyzing the constraints between the assembled components and the upcoming workpiece. The constraints are analyzed in the contact space (column (e)) and converted to the constraint sphere (column (f)) for a qualitative study. We encourage our readers to refer to Wan et al. \cite{wan2018assembly} for details.

An assembly with $n$ workpieces has a sequence of assembly directions for each workpiece represented by $\mathbf{q}$ = ($q_1$ $q_2$ ...  $q_n$), where $q_i$ is the optimal assembly direction of the $i$th workpiece. The sequence of assemblability qualities are $\mathbf{a}$ = ($a_1$ $a_2$ \ldots $a_n$), where $a_i$ is the assemblability quality of the $i$th workpiece. The assemblability quality and assembly directions of $m$ different assembly orders are thus represented as two matrices:
\begin{equation}\label{eq:optimaldirection}
\mathfrak{A}=
\begin{pmatrix}
\mathbf{q_1}\\
\mathbf{q_2}\\
...\\
\mathbf{q_m}
\end{pmatrix} 
= 
\begin{pmatrix}
q_{11} & q_{12} & \ldots & q_{1n} \\
q_{21} & q_{22} & \ldots & q_{2n} \\
\ldots & \ldots & \ldots & \ldots \\
q_{m1} & q_{m2} & \ldots & q_{mn} 
\end{pmatrix}
\end{equation}
and
\begin{equation}\label{eq:assemblyquality}
\mathcal{A}=
\begin{pmatrix}
\mathbf{a_1}\\
\mathbf{a_2}\\
...\\
\mathbf{a_m}
\end{pmatrix} 
= 
\begin{pmatrix}
a_{11} & a_{12} & \ldots & a_{1n} \\
a_{21} & a_{22} & \ldots & a_{2n} \\
\ldots & \ldots & \ldots & \ldots \\
a_{m1} & a_{m2} & \ldots & a_{mn}
\end{pmatrix},
\end{equation}
where $\mathfrak{A}$ is the optimal assembly directions matrix and $\mathcal{A}$ is the assembly quality matrix.

\subsection{Assisting support analyzer}
The assisting support analyzer is triggered when the stability quality of the upcoming workpiece is 0. In that case, the robot is allowed to refer to assisting hands to hold and support the workpiece and maintain balance.

The assisting support analyzer works as follows. Basically, it computes the stability quality of each workpiece sequentially using the same method as the stability analyzer. Through the computation, the analyzer maintains a new set of stability qualities $s_g$ for every workpiece in the finished components. Take an assembly with three workpieces A, B, and C for example. For the order A $\leftarrow$ B $\leftarrow$ C, the analyzer first recomputes the stability of A and set $s_{g1}$ = $s_1$. Then, it recomputes the stability of the complex A and B and set $s_{g2}$ = ($s_1$ $s_2$). In the third step, the analyzer recomputes the stability of the complex made of A, B, and C, and set $s_{g3}$ = ($s_1$ $s_2$ $s_3$). In the process, $s_i$ is updated to $+\infty$ if the original stability quality is 0. The analyzer requires that the number of workpieces with 0 stability quality should be less than the number of avaiable extra hands, for otherwise the robot will not be able to both hold and support the unstable workpiece and at the same time manipulate other workpieces. Note that in the implementation we assume a dual-arm robot and thus the robot can only refer to one extra hand for assistance. In the stability analyzer, the stability quality of a potential order with $n$ workpieces is $\mathbf{s}$=($s_1$ $s_2$ $\dots$ $s_n$). The updated stability quality by the assisting support analyzer is recomputed as $\mathbf{s}$=(min($s_{g1}$) min($s_{g2}$) $\ldots$ min($s_{gn}$)), where $s_{gi}$ is the $i$th group of stability quality. We use the minimum stability quality in the components to represent the new stability quality and exclude the quality of the workpiece already held and supported by an assisting hand. If the group has no other stability quality except the quality from the assisted workpiece, the recomputed stability quality of that group will be positive infinite.

The analyzer suggests the feasible assisting grasps by considering the collisions with the next preparing workpiece (see definitions in part A of this section). It uses the candidate grasps found by the graspability analyzer and reconsiders their collisions with the mesh model of the next preparing workpiece both at its goal state and along its assembly direction. The collision-free grasps will be suggested as the feasible assisting grasps. Note that the grasps are only considered to not block the way of assembling the next preparing workpiece. They may collide with the other hand that grasps the next preparing workpiece. Further collision detection should thus be performed in a consequential motion planner according to the chosen grasp of the next preparing workpiece.

\subsection{Overall evaluation} 
The optimal assembly sequence is chosen by the planner according to evaluating the combination of the stability, graspability, and assemblability in each permutation. 
The optimal solution is found according to the multiplication of the minimum value of each quality, which can be formulated as $min(\mathcal{G})*min(\mathcal{S})*min(\mathcal{A})$. Every row in the matrix represents the overall score of one assembly order. The overall representation of all assembly orders is:
\begin{equation}
\begin{pmatrix}
min(\mathbf{s_1})*min(\mathbf{g_1})*min(\mathbf{a_1}))\\
min(\mathbf{s_2})*min(\mathbf{g_2})*min(\mathbf{a_2}))\\
...\\
min(\mathbf{s_m})*min(\mathbf{g_m})*min(\mathbf{a_m}))
\end{pmatrix} 
\end{equation}
where $\mathbf{s_i}$, $\mathbf{g_i}$, and $\mathbf{a_i}$ correspond to the stability quality, the graspability quality and the assemblability quality in the equations (\ref{eq:stability}), (\ref{eq:graspability}), and (\ref{eq:assemblyquality}) of the $i$th assembly order. Note that the stability quality is from stability analyzer if there exists a stable solution. Otherwise the stability quality is from the assisting support analyzer. The optimal assembly direction can be found by:
\begin{equation}\textit{optimalid}=
argmax_{rowid}
\begin{pmatrix}
min(\mathbf{s_1})*min(\mathbf{g_1})*min(\mathbf{a_1}))\\
min(\mathbf{s_2})*min(\mathbf{g_2})*min(\mathbf{a_2}))\\
...\\
min(\mathbf{s_m})*min(\mathbf{g_m})*min(\mathbf{a_m}))
\end{pmatrix} 
\end{equation}
\begin{equation}
\textit{optimal\_directions} = \mathfrak{A}(optimalid)
\end{equation}

One thing to note it the best solution found by this method is not identical. When the overall indices are the same, the planner will randomly pick one of them as the optimal assembly order.

\section{EXPERIMENT AND ANALYSIS}
\subsection{Assembling three soma blocks}
\begin{figure*}[!htbp]
    \centerline{\includegraphics[width=\textwidth]{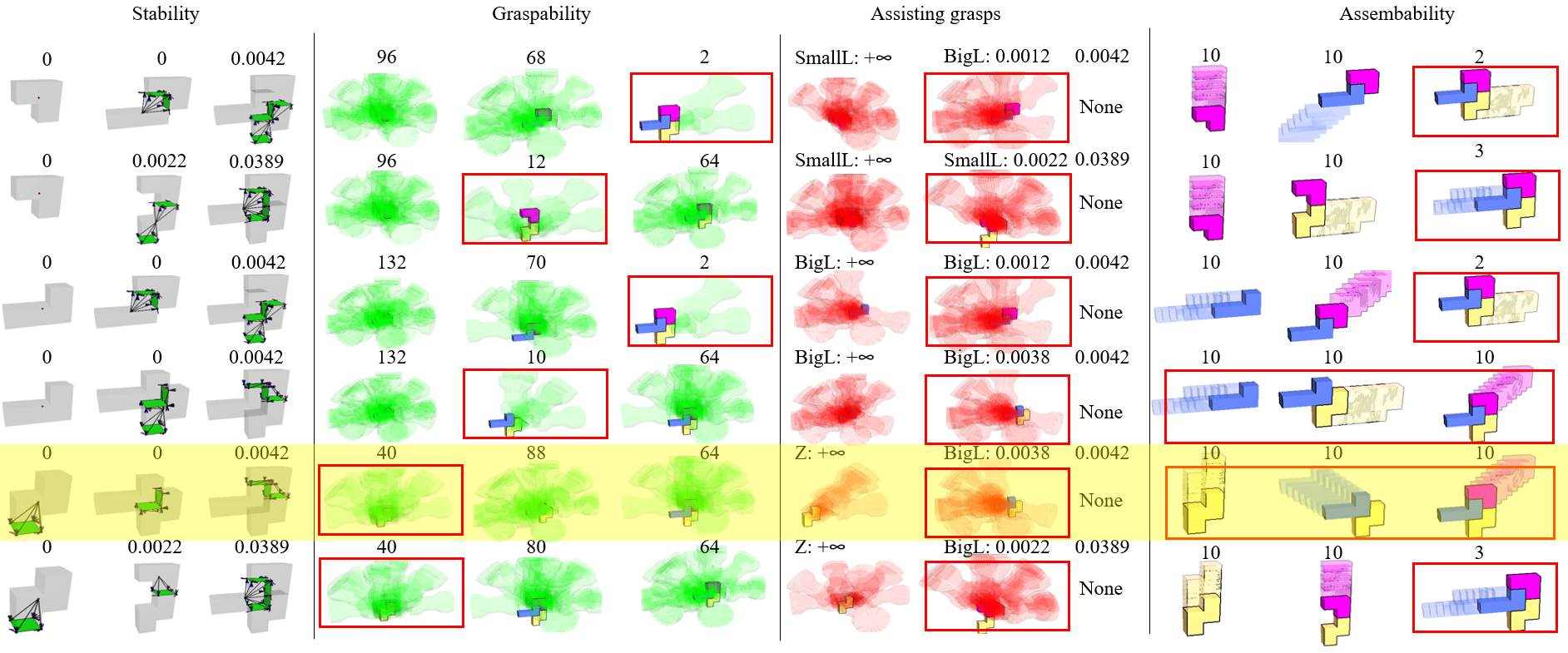}}
    \caption{Each row of the figure is one potential order. The minimum qualities of each order are marked in red boxeds. If all workpieces have the same quality, then the whole order is marked. The optimal assembly sequence is marked by the yellow banner.}
    \label{fig:ex1}
\end{figure*}
An example of a soma block assembly using the input shown in Fig.\ref{fig:intro} is presented in Fig.\ref{fig:ex1}. Here, we name the blue workpiece ``BigL'', the pink workpiece ``smallL'', and the yellow workpiece ``Z''. The example has six possible assembly orders. All of them contain unstable workpieces in the middle, but their final assembly state is stable. The four columns separated by the solid line in the figure correspond to the stability, graspability, assisting grasps, and the assemblability computed using the method presented in Section IV. In the stability column, each row shows the results of one potential assembly order. The index on the top of each image in the column is the stability quality $\mathcal{S}$. In the graspability column, the green hands show the candidate grasps for each upcoming workpieces. Likewise, the index on the top of each small figure indicates the graspability quality $\mathcal{G}$. The assisting grasps column shows the unstable workpieces and the assisting grasping poses. The red hands are the assisting grasping poses. The unstable workpieces are blocked by the hands. To make them clear, we additionally placed the workpiece name and recomputed stability index on top of each small figure to show the details. None of the first workpieces is stable and they have to be held by an assisting grasp. Thus, there are always red hands for them. Also, the stability quality is always +$\infty$ since there is only one workpiece in the assembled component. The stability of the second workpiece is the workpiece that has the smallest stability in the assembled components. They are therefore no longer infinite values. The third workpiece is at the final stable state and there is no need for assisting grasps. We simply dropped a ``None'' there to clarify this state. The index in the assemblability column represents the assemblability quality $\mathcal{A}$. Since all the assembly orders contain unstable workpieces, we evaluate the overall best order and directions by only considering the stability quality recomputed by the assisting support analyzer. The optimal assembly order is the second to the last row (labeled in yellow color). Readers may inspect each of the columns for the detailed accessible grasps, assisting grasps, and assembly directions of this optimal result.

\subsection{Other results}
\begin{figure}[!htbp]
    \centerline{\includegraphics[width=\linewidth]{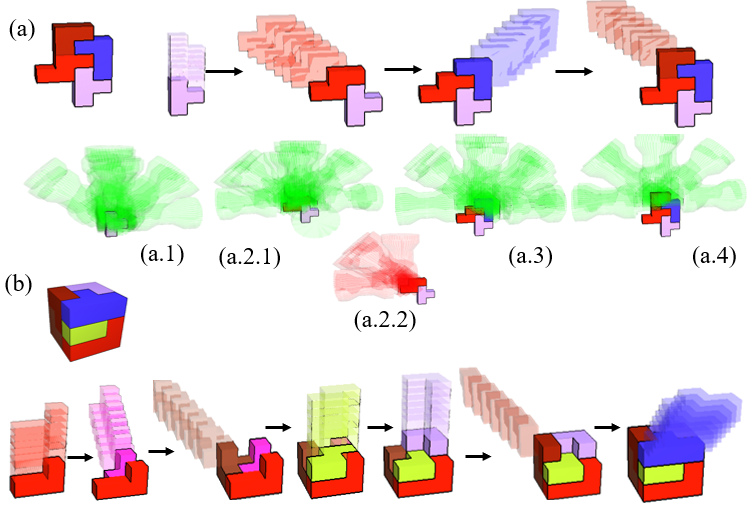}}
    \caption{(a) Planning to assemble four soma workpieces. Assembling the ``Z'' workpiece in (a.2.2) requires assisting grasps.
    (b) Planning to assemble seven soma workpieces. No assiting grasps are needed in this case.}
    \label{fig:ex2}
\end{figure}

The Fig.\ref{fig:ex2} shows some other planning results generated by the proposed planner. Fig.\ref{fig:ex2}(a) is a four-workpiece assembly which involves unstable workpieces in every possible order. The first row is the optimal assembly order and the assembly direction of each component. The second row is the grasps and assisting grasps of the assembly. The grasps in green color are the grasps for the workpiece and the grasps in red are the assisting grasps. Here, assembling the ``Z'' workpiece in (a.2.2) is unstable so that assisting grasps are included. After assembling the "BigL" workpiece, the assembly is restabilized and the assisting grasp is no longer needed. The Fig.\ref{fig:ex2}(b) is an example of 7 workpieces assembling into a cube. Our planner successfully found a stable assembly sequence without any reference to assisting grasps. The final assembly order and assembly directions are shown in the picture.

\subsection{Real-world execution}
The planned result can be sent to a motion planner (see \cite{wan2016achieving}) for generating the robot motions to perform the assembly task. Fig.\ref{fig:real} shows an examples. The detailed videos are in the supplementary file. During execution, F/T (force and torque) sensors are used to make the pick-and-place stable. 
\begin{figure*}[!htbp]
    \centerline{\includegraphics[width=\textwidth]{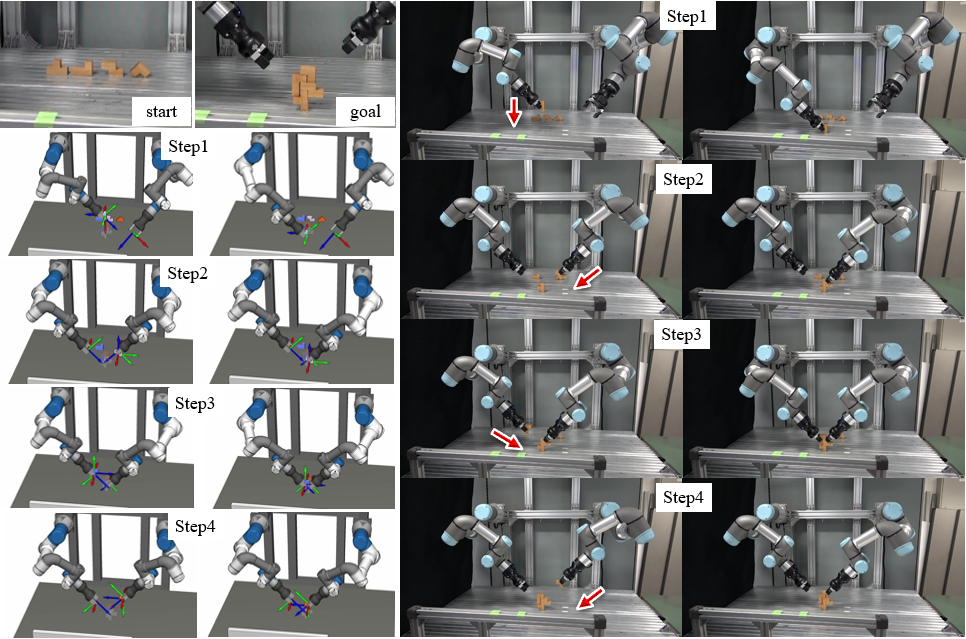}}
    \caption{Top-left: The start and goal state of the soma blocks. Left: Simulation results. Right: The robot assembles the workpieces according to the sequence, directions, grasping poses, and assisting grasps found by the proposed planner. The red arrows indicate the assembly direction. }
    \label{fig:real}
\end{figure*}

\section{CONCLUSIONS}

An assembly planner is proposed in this paper to plan an optimal assembly sequence. The planner can handle the unstable workpiece during assembly by using the assisting grasps to hold and support the unstable workpiece. The planned sequences can be used by a motion planner to plan the motion for dual-arm assembly. Currently, the planner is established for the soma block. In the future, we will adapt the proposed planner to objects with non-trivial shapes and make the algorithms general.

\bibliographystyle{IEEEtran}
\bibliography{IEEEabrv,reference}

\end{document}